\documentclass[10pt,twocolumn,letterpaper]{article}
\newcommand\blfootnote[1]{%
  \begingroup
  \renewcommand\thefootnote{}\footnote{#1}%
  \addtocounter{footnote}{-1}%
  \endgroup
}
\usepackage{cvpr}              % To produce the CAMERA-READY version
\usepackage[dvipsnames]{xcolor}

\definecolor{cvprblue}{rgb}{0.21,0.49,0.74}
\usepackage[pagebackref,breaklinks,colorlinks,citecolor=cvprblue]{hyperref}
\def\paperID{*****} % *** Enter the Paper ID here
\def\confName{CVPR}
\def\confYear{2024}

\title{CityLLaVA: Efficient Fine-Tuning for VLMs in City Scenario}

\author{
    Zhizhao Duan$^{{*}}$ \quad
    Hao Cheng$^{{*}}$ \quad
    Duo Xu \quad
    Xi Wu \quad
    Xiangxie Zhang \quad
    Xi Ye \quad 
    Zhen Xie$^\dagger$ \quad
    \vspace{0.5em}\\
Alibaba Group \\
{\tt\small \{zhizhao.dzz,luchen.ch,manii.xd,lingke.wx\}@alibaba-inc.com} \\
{\tt\small \{zhangxiangxie.zxx,yx150449,xiezhen.xz\}@alibaba-inc.com}
}

\begin{document}
\maketitle
{\blfootnote{* Authors contributed equally to this work.}}
{\blfootnote{\textdagger \ Corresponding authors.}}
\begin{abstract}

In the vast and dynamic landscape of urban settings, Traffic Safety Description and Analysis plays a pivotal role in applications ranging from insurance inspection to accident prevention. This paper introduces CityLLaVA, a novel fine-tuning framework for Visual Language Models (VLMs) designed for urban scenarios. CityLLaVA enhances model comprehension and prediction accuracy through (1) employing bounding boxes for optimal visual data preprocessing, including video best-view selection and visual prompt engineering during both training and testing phases; (2) constructing concise Question-Answer sequences and designing textual prompts to refine instruction comprehension; (3) implementing block expansion to fine-tune large VLMs efficiently; and (4) advancing prediction accuracy via a unique sequential questioning-based prediction augmentation. Demonstrating top-tier performance, our method achieved a benchmark score of 33.4308, securing the leading position on the leaderboard. The code can be found: \href{https://github.com/alibaba/AICITY2024_Track2_AliOpenTrek_CityLLaVA}{https://github.com/alibaba/AICITY2024\_Track2\_\\AliOpenTrek\_CityLLaVA}.

\end{abstract}

% (1) Use the area of the bounding boxes to select the good view of video for training and testing; (2) Use the bounding boxes to crop the ROI region and build  global and local views with visual prompts to enhance the  visual embeddings;    
\section{Introduction}
\label{sec:intro}

With the rapid advancement of large language models (LLMs), an increasing number of fields are beginning to explore the capabilities of these models, investigating their potential impact on industry standards and societal practices. Particularly in research areas that straddle computer vision (CV) and natural language processing (NLP), such as traffic video analysis, these models have not only significantly raised the bar for automated analysis precision but have also unlocked unprecedented vistas of application. There are myriad foundational visual-language models (VLMs) such as GPT4-V~\cite{2023GPT4VisionSC}, Qwen-VL-Chat~\cite{qwen}, LLaVA~\cite{liu2023llava} and others that stand as testaments to the synergetic potential of CV and NLP. While these large models exhibit formidable capabilities across a range of tasks, they often fall short of expectations when applied directly to highly specialized domains, such as traffic safety scenario captioning. It becomes evident that these models require essential fine-tuning to adequately capture the domain-specific nuances.
\begin{figure}[t]
    \centering
    \includegraphics[width=0.5\textwidth]{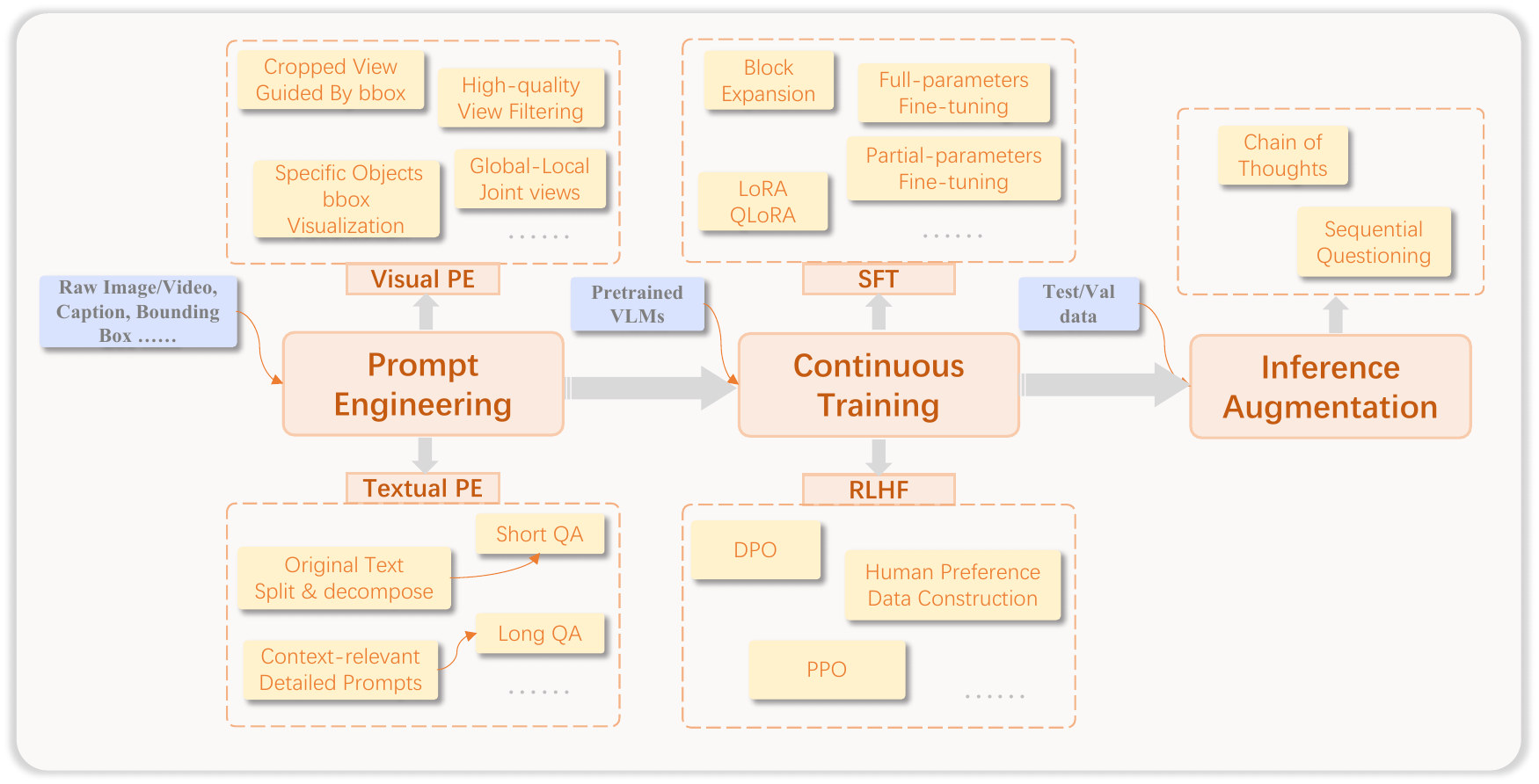}
    \caption{The efficient fine-tuning paradigm for VLMs. The efficient fine-tuning paradigm for VLMs. The paradigm first executes the prompt engineering, which includes visual and textual prompt engineering. Then continuous training including SFT and RLHF is implemented based on the pretrained VLMs. Finally, the Inference augmentation is used to improve performance.}
    \label{fig: paradigm}
\end{figure}

Fine-tuning a large model to meet the granular needs of a particular application involves navigating a complex landscape of challenges, such as the construction of effective question-answer pairs, the design of appropriate prompts, and the selection of critical fine-tuning parameters. Furthermore, creating annotations that capture the multifaceted nature of real-world events can be labor-intensive.

In light of these challenges, this work aims to introduce an effective and comprehensive paradigm for fine-tuning large visual-language models, including prompt engineering, continuous training, and inference augmentation.  Figure~\ref{fig: paradigm} shows the details of this effective paradigm, which is accumulated in our industry practice. To demonstrate the effectiveness of the proposed approach, we conduct detailed experiments on the WTS~\cite{WTS2024} dataset. Specifically, the proposed paradigm has achieved first place in the 2024 AI City Challenge~\cite{Shuo24AIC24} - Traffic Safety Description and Analysis, contributing valuable insights and offering a pathway for future research to enhance or adapt LLMs for similar complex, domain-specific tasks. 

In sum, the contributions of this paper are summarized as follows: 
\begin{itemize}
    \item Proposing an effective and comprehensive paradigm for fine-tuning large visual-language models for domain-specific tasks.
    \item Exploring visual and textual prompt engineering to construct informative and refined inputs for both training and inference.
    \item Investigating the adaptation of block expansion in VLMs, achieving superior performance compared to LoRA~\cite{hu2022lora}.
    \item Achieving state-of-the-art performance on the WTS dataset, with a detailed exploration of factors influencing fine-tuning efficacy.
\end{itemize}
\section{Related Work}
\label{sec:related_works}

\begin{figure*}[htb]
    \centering
    \includegraphics[width=\textwidth]{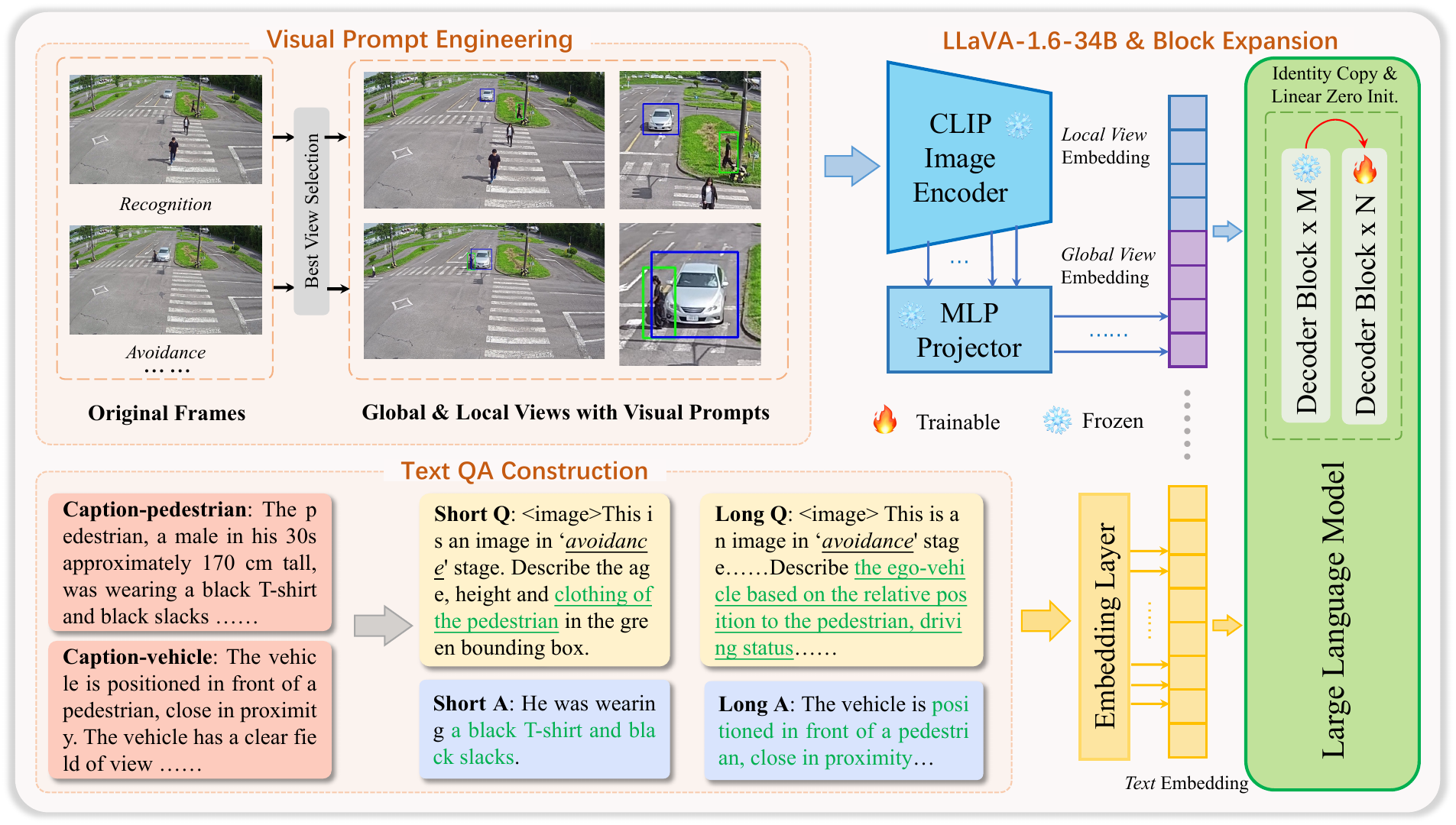} 
    \caption{The overview of CityLLaVA. Our method is anchored on the pretrained LLaVA-1.6-34B~\cite{liu2024llavanext} equipped with block expansion~\cite{wu2024llama}, combining the textual prompt engineering and visual prompt engineering guided by bounding boxes.} 
    \label{fig:arch_overview} 
\end{figure*}

\subsection{Vision-Language Models}

Vision-language models (VLMs) utilize image and text data simultaneously and fuse knowledge in different domains for better performance. CLIP~\cite{radford2021learning} is a pioneering work that aligns language and image by designing a pretraining task that matches images with text captions. It shows spectacular zero-shot transferability among several downstream tasks. In recent years, with the development of large language models~\cite{ouyang2022training,touvron2023llama,qwen}, combining a visual encoder with an auto-regressive language decoder has become a prevalent approach in vision-language tasks. This type of method can benefit from both visual perception and linguistic expression and a more versatile model can be realized. An early study in this area is Flamingo~\cite{alayrac2022flamingo}, which leverages gated cross-attention to accept interleaved visual and language data as input and then generates text as output. BLIP-2~\cite{li2023blip} introduces a lightweight but powerful module named Q-former to efficiently bridge the modality gap between image and text, while FlanT5~\cite{chung2022scaling} is used as the language model. Built on the pretrained visual component of BLIP-2, Mini-GPT4~\cite{zhu2023minigpt} employs a single projection layer to align the visual features with text features and input to the Vicuna~\cite{vicuna2023} language model. An improved version is MiniGPT-v2~\cite{chen2023minigptv2}, which applies a simpler strategy that directly projects the visual tokens from a ViT~\cite{dosovitskiy2020vit} encoder to the feature space of a large language decoder. LLaVA~\cite{liu2023llava} adopts a similar model structure of utilizing a projection layer after the encoded visual features. With the proposed two-stage training strategy, LLaVA demonstrates impressive abilities in vision-language tasks and there are many following works based on it~\cite{liu2023improvedllava, liu2024llavanext, sun2023aligning, lin2023video, luo2024feast}.

\subsection{VLMs in Driving}
Many researchers have attempted to apply vision-language models in driving since they have shown remarkable capabilities in visual signal perception and language understanding. A previous study~\cite{wen2023road} conducts an exhaustive evaluation on the state-of-the-art vision-language model GPT4-V~\cite{2023GPT4VisionSC} in the autonomous driving scenario and the experiment results illustrate superior performance. Dolphins~\cite{ma2023dolphins}, a novel vision-language model in which the pretrained OpenFlamingo~\cite{alayrac2022flamingo} serves as the fundamental structure, demonstrates distinctive behaviors in the driving domain. DriveGPT4~\cite{xu2024drivegpt4} can process textual queries and multi-frame videos as the input and generate corresponding responses, while it is also capable of predicting low-level vehicle control actions and signals. Experiment results suggest that DriveGPT4 has comparable or even better ability in some cases compared with GPT4-V. 
 
\label{subsec:}
\section{Methodology}
\label{sec:methodology}

\begin{figure*}[htb]
    \centering
    \includegraphics[width=\textwidth]{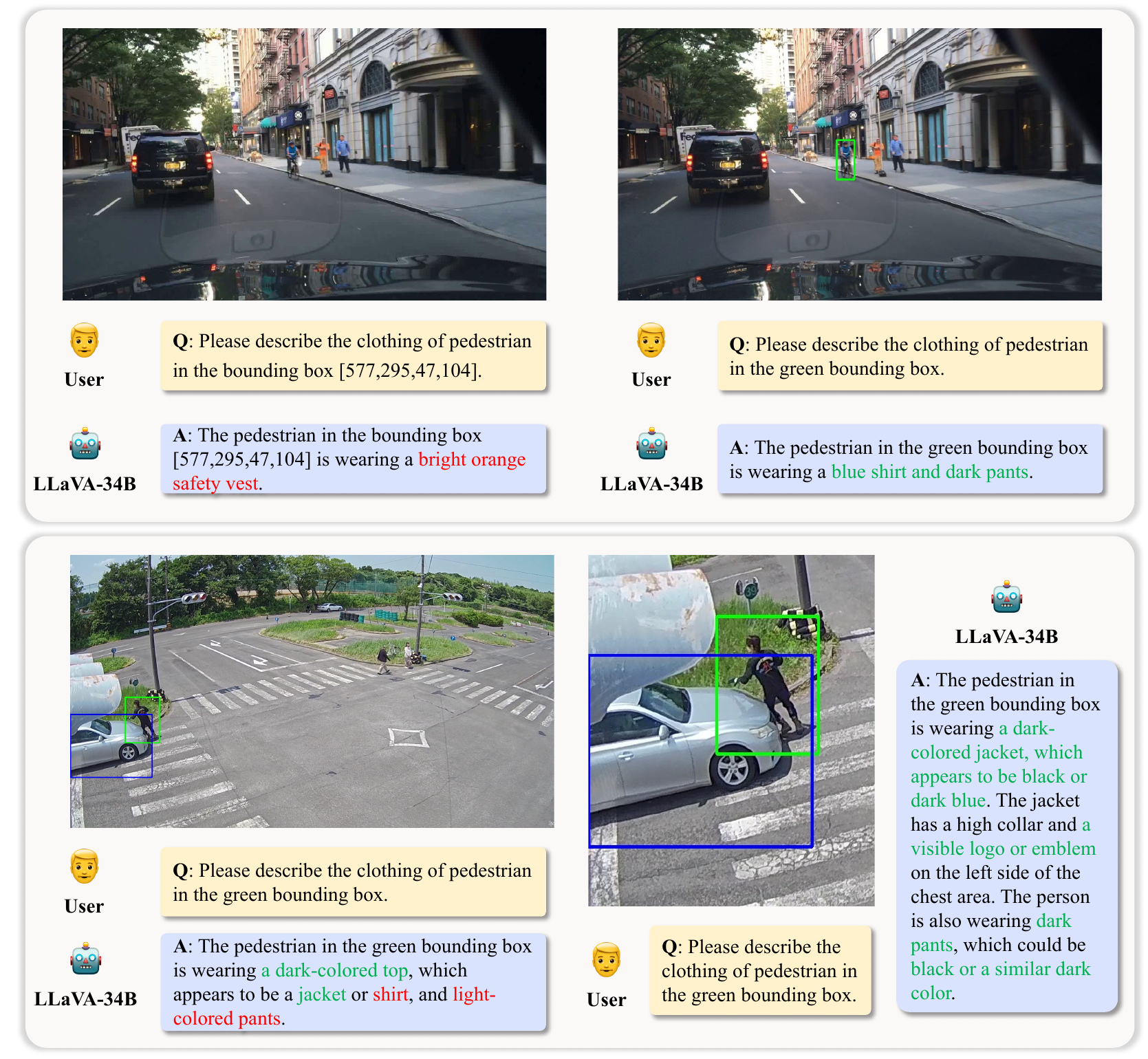} 
    \caption{Examples of usages of visual prompt (Top) and cropped view guided by bounding boxes (Bottom).} 
    \label{fig:effectiveness_visual_prompt} 
\end{figure*}

\subsection{Overview}

CityLLaVA introduces an efficient fine-tuning pipeline aimed at enhancing spatial-temporal understanding and providing fine-grained perception within urban environments. As shown in Figure~\ref{fig:arch_overview}, the proposed paradigm consists of three major modules: visual prompt engineering, textual prompt engineering (i.e., text QA construction), and efficient fine-tuning for the large vision-language model. We will sequentially introduce the details of each module.

\subsection{Dataset Construction}
In this section, we elaborate on the details of data filtering and the construction of a vision-language instruction-tuning dataset for Supervised Fine-tuning (SFT). For the fine-tuning, an item $\mathbf{H}$ for training can be formulated as a tuple:
\begin{equation}
    \mathbf{H} = (\mathbf{X_v}, \mathbf{X_q}, \mathbf{X_t})
\end{equation}
where $\mathbf{X_v}, \mathbf{X_q}, \mathbf{X_t}$ denote the visual inputs, textual instruction and textual response, respectively. The best construction of them is explored in the subsequent sections.

\subsubsection{Bounding-box Guided View Selection}
The WTS dataset consists of two parts: a. WTS data, which is a multi-view dataset with an uncertain number of vehicle views and overhead views; b. Filtered pedestrian-centric videos from BDD100K~\cite{bdd100k} data with vehicle view only. Given that the relevant vehicles and pedestrians might not be clearly visible or could be insignificant in size within some views of the WTS data, directly fine-tuning Visual Language Models (VLMs) on multi-perspective data poses a challenge. To address this, we introduce bounding-box guided view selection. Initially, we filter out overhead views from the WTS data that do not match the officially recommended perspectives. The data is then segmented into tuples $\mathbf{S}=\{\mathbf{V}, \mathbf{T}, \mathbf{B}\}$, representing the video clips, descriptions, and bounding boxes, following the segment set forth by the authorities. For each tuple $\mathbf{S}_i$ in the training dataset, we compute the average vehicle area $\mathbf{A_v}_i$ and the average pedestrian area $\mathbf{A_p}_i$ based on the bounding boxes $\mathbf{B}_i$. We consider the area to be zero if a bounding box is not present. We then apply the following criteria to obtain the filtered tuple data $\hat{\mathbf{S}}_i$, ensuring that only relevant views are selected for model training:
% \hat{S}=Filter\left(S_i\right)=\left\{\begin{matrix}\begin{matrix}False&A_{p_i}>{thr}_p,\ {\ A}_{v_i}>{thr}_v\\False&A_{p_i}>{thr}_p\ &\ A_{v_i}=0\\\end{matrix}\\\begin{matrix}False&A_{v_i}>{thr}_v\ \&\ A_{p_i}=0\\True&otherwise\\\end{matrix}\\\end{matrix}\right.
\begin{equation}
\hat{\mathbf{S}}_i = 
\begin{cases}
    \mathrm{False}, & \text{if } A_{p_i} > {thr}_p \text{ and } A_{v_i} > {thr}_v, \\
    \mathrm{False}, & \text{if } A_{p_i} > {thr}_p \text{ and } A_{v_i} = 0, \\
    \mathrm{False}, & \text{if } A_{v_i} > {thr}_v \text{ and } A_{p_i} = 0, \\
    \mathrm{True}, & \text{otherwise}.
\end{cases}
\end{equation}
where $\mathrm{thr}_p$ and $\mathrm{thr}_v$ denotes the area threshold of pedestrian and vehicle.

For the $\texttt{test}$ dataset, we initially group the tuples $\mathbf{S}$ by scenario. Within a scenario, we select the video that not only features bounding boxes across five stages but also showcases the largest average pedestrian area $\mathbf{A_p}$  as the optimal view for testing purposes. This process ensures the selection of the most suitable perspective for inference.

\subsubsection{Visual Prompt Engineering}
\label{subsec:vpe}
In this section, the visual prompt engineering (VPE) is elaborated from two aspects. 

\noindent\textbf{Visual Prompt}. A visual prompt introduces an innovative approach to prompt engineering. VLMs are inherently multimodal and provide the chance to manipulate both visual and textual modality. A simple red circle in pixel space can direct CLIP’s attention to the interested region improving performance in zero-shot referring expressions comprehension~\cite{shtedritski2023does}. ViP-LLava~\cite{cai2023vipllava} also leverages a red bounding box or point arrow to enhance the region-level perception of VLM. In this paper, we introduce bounding box rectangles as visual prompts to localize the interested pedestrian and vehicle, for the sake of fine-grained vision-language alignment and information extraction of local region.

We implement visual prompts by drawing the scaled-up bounding boxes on the corresponding video frame. Given a frame $\mathbf{I}$, and its pedestrian bounding box $\mathbf{B}_{p}$, vehicle bounding box $\mathbf{B}_{v}$, we visualize the bounding box with green rectangle for the pedestrian and blue rectangle for the vehicle. It is critical to scale up the bounding box to ensure it encompasses the entire region of interest around the pedestrian or vehicle, as the original bounding box might cover only a partial area.  Given a bounding box $\mathbf{B}$, the scaled-up $\hat{\mathbf{B}}$ can be formulated as the follow:
\begin{equation}
    \hat{\mathbf{B}} = \text{Scale}(\mathbf{B}, c) = (x + \frac{w}{2} - \frac{cw}{2}, y + \frac{h}{2} - \frac{ch}{2}, cw, ch)
    \label{eq:scale_up_bbox}
\end{equation}
where $c$ denotes the scaling coefficient, $x$ and $y$ represent the coordinates of the top-left corner, and $w$ and $h$ represent the width and height of the box, respectively. The default $c$ is set to $1.2$. Based on the scaled-up bounding boxes $\hat{\mathbf{B}}_p=(\hat{x}_p,\hat{y}_p, \hat{w}_p, \hat{h}_p)$ and $\hat{\mathbf{B}}_v=(\hat{x}_v,\hat{y}_v, \hat{w}_v, \hat{h}_v)$, we can generate the augmented frame $\hat{\mathbf{I}}$ with visual prompts, which is then fed into VLM.

\noindent\textbf{Global-Local Joint Views}. To enhance the understanding of both global contexts and specific interests in pedestrians and vehicles, we concatenate the augmented frame $\hat{\mathbf{I}}$ and its cropped view guided by bounding boxes $\hat{\mathbf{I}}_m$ as joint visual inputs. Prior research~\cite{lin2023sphinx, liu2024llavanext} suggests that using a concatenation of multi-cropped and full-view images as visual inputs can enhance the performance on fine-grained multi-modal understanding and reduce the hallucination in outputs. However, this manner brings a greater computational burden. Furthermore, the visual redundancy introduced in this procedure may yield fewer improvements, even decrements for certain tasks. As a result, We replace multi-cropped views in~\cite{liu2024llavanext} with local cropped views guided by bounding boxes. This replacement directs the model's attention to the key region while reducing unnecessary visual information. Given bounding boxes $\hat{\mathbf{B}}_p$ and $\hat{\mathbf{B}}_v$, the cropped view boundary can be formulated as:
\begin{equation}
\begin{gathered}
    \begin{split}
        \mathbf{V}_{m} &= \text{Crop}(\hat{\mathbf{B}}_p, \hat{\mathbf{B}}_v) \\
                       &= (x_{\min}, y_{\min}, x_{\max}-x_{\min}, y_{\max}-y_{\min})
    \end{split} \\
    x_{\min} = \min(\hat{x}_p, \hat{x}_v) \\
    y_{\min} = \min(\hat{y}_p, \hat{y}_v) \\
    x_{\max} = \max(\hat{x}_p + \hat{w}_p, \hat{x}_v + \hat{w}_v) \\
    y_{\max} = \max(\hat{y}_p + \hat{h}_p, \hat{y}_v + \hat{h}_v)
\end{gathered}
\end{equation}
where $\mathbf{V}_{c}$ denotes the boundary of the cropped view, which can be interpreted as the smallest external rectangle of two bounding boxes. We can use $\text{Scale}(\cdot, \cdot)$ defined in Eq.~\ref{eq:scale_up_bbox} to scale up the boundary for more context attributes:
\begin{equation}
    \hat{\mathbf{V}}_{m}=\text{Scale}(\mathbf{V}_m, c^*)
\end{equation}
Here $c^*$ is set to 1.5. The final cropped view $\hat{\mathbf{I}}_m$ can be formulated as:
\begin{equation}
    \hat{\mathbf{I}}_m = \hat{\mathbf{I}}[:,\hat{x}_m:\hat{x}_m+\hat{w}_m, \hat{y}_m:\hat{y}_m+\hat{h}_m]
\end{equation}
where $\hat{x}_m, \hat{y}_m$ represent the coordinates of top-left corner of $\hat{\mathbf{V}}_{m}$, and $\hat{w}_m, \hat{h}_m$ denote the corresponding width and height. Both $\hat{\mathbf{I}}_m$ and $\hat{\mathbf{I}}$ are fed into the vision encoder during training and inference.

\noindent\textbf{Effectiveness of VPE}. We conduct a simple experiment to verify the effectiveness of our visual prompt engineering. We use the textual prompt "\textit{Please describe the clothing of pedestrian \{in the bounding box [x, y, h, w]\}/\{in the green bounding box\}}" to query the original LLaVA-1.6-34B~\cite{liu2024llavanext} under the original frames and augmented or cropped frames, and compare the accuracy of outputs in different settings. As shown in Figure~\ref{fig:effectiveness_visual_prompt}, the proposed paradigm enhances the fine-grained perception in specific objects and interested regions. The sub-figure on the top illustrates that the drawn green rectangle directs the model's attention to the interested pedestrian who is more likely to be collided with the ego-vehicle. These simple visual prompts prevent the model from being disturbed by irrelevant objects, building up the fine-grained visual-language alignment. The sub-figure on the bottom demonstrates the improvement in detail recognition during inference with a cropped view guided by bounding boxes. The experiments conducted indicate that datasets processed with the proposed visual prompt engineering exhibit enhanced alignment, which is conducive to the fine-tuning of the model.

\subsubsection{Textual Prompt Engineering}
\label{sec:text_prompt_engineering}
A well-constructed prompt is essential for visual question answering and visual captioning, especially when detail-intensive descriptions are required. Through prompt engineering, we aspire to identify a question that not only encapsulates the content of the description accurately but also comprehensively, rather than just inputting "Please provide a detailed description of the pedestrian/vehicle in the video" into the model. 

By conducting a dimensional analysis of the descriptions, we distill the key points such as height, clothing, line of sight, relative position, movement, and environment. We engage GPT-4v \cite{2023GPT4VisionSC} to evaluate the alignment of the generated responses with ground truth, recognizing mismatches and areas for enhancement. The result is a set of prompts that are meticulously crafted to guide the model towards generating high-quality, context-relevant answers.

\noindent\textbf{Prompt for pedestrian descriptions. } This picture shows the relationship between the pedestrian in the green box and the vehicle in the blue box. Describe the pedestrian in the green box or the pedestrian closest to the vehicle based on age, height, clothing, line of sight, relative position to the vehicle, movement status, weather conditions, and road environment.

\noindent\textbf{Prompt for vehicle descriptions. } This picture shows the relationship between the vehicle in the blue box and the pedestrian in the green box. Describe the vehicle in the blue box or the vehicle closest to the pedestrian based on the relative position to the pedestrian, driving status, weather conditions, and road environment. And describe the age, height, and clothing of the pedestrian.

For the vehicle perspective, we utilize "ego-vehicle" in place of "the vehicle in the blue box" to enhance the contextual relevance and specificity of the prompt.

\subsubsection{Short QA Construction}
\label{sec:short_qa_construction}
Verbose descriptions often hinder a model's alignment with pertinent content (e.g., \textit{localization}, \textit{attention}, \textit{context attributes}). To tackle this issue, we have introduced a series of short question-answer (QA) pairs derived from the source descriptions to enhance dataset diversity. This approach aims to reduce model output style and template over-fitting during the fine-tuning process. By splitting each description into specific dimensions - attributes, location, motion state, and environment -  we can construct targeted questions that elicit detailed and relevant responses from the model.

To ensure the preservation of the context and structure with the generated data, we develop a description splitting method that utilizes GPT-4~\cite{2023GPT4} to categorize each sentence of the descriptions into predefined dimensions. The sentences within the same dimension are concatenated to form a cohesive segment. This segment is then paired with the corresponding query for that dimension to construct a tailored question-answer pair. The prompt used for this process is as follows: "Please select the most appropriate label for each descriptive text from the following options, and format the output by providing the text index followed by the letter a, b, c, d, or e. Each selection should be on a new line."

Finally, we construct an image-text dataset including long QA and short QA pairs. Note that there are two different compositions of textual parts: (1) \textit{Multi-round QA}, a multi-round conversation including both long QA defined in Sec.~\ref{sec:text_prompt_engineering} and short QA pairs. (2) \textit{Single-round QA}, a single-round conversation including just a QA pair. We compare the influence of these two manners in Sec.~\ref{sec:ablation_study}

\subsection{Model Architecture}

Initially, Qwen-VL-Chat~\cite{qwen}and Video-LLaVA~\cite{lin2023video}  are selected as the candidate baseline model for this video understanding task due to their relatively good performance. Both of these models extract 8 frames uniformly from each stage video clip ${\mathbf{V}}$ as input during the data processing. Concerning the fine-tuning approach, inspired by LLaMA-Pro~\cite{wu2024llama}, we use the block expansion instead of LoRA to fine-tune the baseline model. In the block-expansion method, some zero linear Initialized decoder block layers, which are identity copied from the LLM module of VLM, are interleaved into the LLM backbone. During fine-tuning, only the parameters of these duplicate block layers are unfrozen. This method demonstrated enhanced learning capabilities, resulting in improved performance indicators. A detailed comparison between LoRA and block expansion can be found in Table~\ref{tab:backbone_sft}. The result reveals that Qwen-VL-Chat significantly outperforms Video-LLaVA. 

However, it was observed that many stage video clips contain fewer than 8 frames, sometimes only one.  It implies that we can deal with this task with an image model. Considering the first frame of each video clip  ${\mathbf{V}}$ is manually annotated, and the annotation quality surpassed that of the remnant tracking data, we deploy the LLaVA-1.6-34B~\cite{liu2024llavanext}, which is the state-of-the-art VLM. We only use the first frame of each video clip  ${\mathbf{V}}$ as the input and also apply the block expansion for efficient fine-tuning. Table~\ref{tab:model_param} shows the number of model's parameter with block-expansion. We find that despite the LLaVA-1.6-34B with single frame input losing some temporal information possibly, its larger parameter size aided in a more fine-grained understanding of the image.
\begin{table}[]
    \centering
    \resizebox{0.4\textwidth}{!}{
        \begin{tabular}{c|cccc}
        \hline
            Model& Blocks& Dim& Heads& Parameters\\
        \hline
        Qwen-VL-Chat& 32& 4096& 32& 7B\\
        Video-LLaVA& 32& 4096& 32& 7B\\
        LLaVA-34B& 60& 7168& 56& 34B\\
 \hline
 Qwen-VL-Chat+BE& 40& 4096& 32&9B\\
 Video-LLaVA+BE& 40& 4096& 32&9B\\
 LLaVA-34B+BE& 72& 7168& 56&41B\\
 \hline
        \end{tabular}
    }
    \caption{the model's parameter w/o block-expansion. \textbf{BE} refers to the block expansion~\cite{wu2024llama}.}
    \label{tab:model_param}
\end{table}

Furthermore, we attempted to enhance the model's performance by implementing Reinforcement Learning from Human Feedback (RLHF). Previous work~\cite{sun2023aligning} shows that LLaVA can benefit from RLHF by reducing hallucinations. However, the original RLHF is based on the PPO algorithm~\cite{schulman2017proximal}, which consumes large computational resources. We choose the Direct Preference Optimization (DPO) algorithm~\cite{rafailov2024direct} as an alternative since it is more computationally efficient. Following previous work~\cite{yu2023rlhf, zhao2024hallucinations}, our target is using DPO to alleviate hallucinations and improve performance. We leverage the ground truth of the phase descriptions as positive samples and the outputs from the SFT model as negative samples, then feed them into the DPO. Unfortunately, DPO leads to a decline in performance indicators. Upon analysis, we find two possible reasons for the degraded performance. Firstly, the description in this task is relatively longer than the captions or responses in other datasets, making DPO harder to align. Secondly, various annotation templates have been applied in the dataset, which might confuse DPO such that the model does not know which template needs to be aligned. Therefore, we remove DPO training from this challenge.

\subsection{Harnessing Sequential Questioning}

We have investigated the impact of sequential questioning on the performance of the model trained exclusively on single-round QA instances. Despite the absence of multi-round QA pairs in the training dataset (i.e., the items in \texttt{train} set have no sequential question-answer pairs containing both of the "\textit{pedestrian}" prompts and "\textit{vehicle}" prompts defined in Sec.~\ref{sec:text_prompt_engineering} at the same time), our findings reveal an improvement in response accuracy when the model is subjected to a series of questions in a specific order during inference.

A general phenomenon has been discovered that "\textit{vehicle}" scores are higher than "\textit{pedestrian}" scores on average. This comparison indicates that the model's output for the "\textit{vehicle}" prompt contains a more precise description of context attributes, localization, and attention. Therefore, a reasonable approach is to insert the "\textit{vehicle}" description into the "\textit{pedestrian}" prompt providing enhanced contexts for a more precise "\textit{pedestrian}" description. The above strategy can be summarized into "Firstly asking \textit{vehicle}, then asking \textit{pedestrian}."

By simply leveraging the sequential questioning, the model conducts outputs with higher evaluation metrics, which can be regarded as a prediction augmentation. The detailed experimental results for the sequence of the questions are analyzed in Table~\ref{tab:sequential_qa}. Note that only a specific order of questions can improve model performance.
\begin{table}[t]
    \centering
    \resizebox{0.4\textwidth}{!}{
        \begin{tabular}{c|c|c|c}
            \hline
            Rank & Team ID & Team Name & Score \\
            \hline
            1 & 208 & \textbf{AliOpenTrek (ours)} & \textbf{33.4308} \\
            2 & 28 & AIO\_ISC & 32.8877 \\
            3 & 68 & Lighthouse & 32.3006 \\
            4 & 87 & VAl & 32.2778 \\
            5 & 184 & Santa Claude & 29.7838 \\
            6 & 34 & LTDT & 29.4070 \\
            \hline
        \end{tabular}
    }
    \caption{\textbf{Leaderboard of Traffic Safety Description and Analysis}. Our method ranks first place in the 2024 AI City Challenge Track 2.}
    \label{table: ranking}
\end{table}

\section{Experiments}
In this section, we explain the datasets and metrics firstly. Subsequently, we introduce our implementation details. We also provide the results on the ablation study.
\subsection{Dataset}
This paper uses the WTS dataset for the model training and evaluation. WTS is the largest dataset for spatial-temporal fine-grained video understanding in the traffic domain, aiming at describing detailed behaviors of both vehicles and pedestrians within a variety of staged traffic events including accidents. WTS features over 1,200 video events from more than 130 distinct traffic scenarios, combining perspectives from both ego-vehicle and fixed overhead cameras within a vehicle-infrastructure cooperative environment. It offers detailed textual descriptions for each event, covering observed behaviors and contexts. Additionally, for broader research applications, detailed textual annotations are also available for 4,861 publicly accessible pedestrian-centric traffic videos from BDD100K.

Because of the large number of \texttt{val} set in the WTS dataset and the large computational resources usage, we conduct the ablation experiments on a selected subset of the original \texttt{val} set, which contains 82 samples, a total 301 entries.

\subsection{Evaluation Metrics}
For CityLLaVA, we use BLEU-4, METEOR, ROUGE-L, and CIDEr as evaluation indicators to compare the predicted descriptions against the ground truth. More specifically, these 4 metrics are used to calculate a final score:
{\small\begin{equation}
    \text{Score} = \frac{\text{BLEU-4} + \text{METEOR} + \text{ROUGE-L} + 0.1\times\text{CIDEr}}{4} \times 100
    \label{eq:metric}
\end{equation}}
% \begin{table}[h]
% \centering
% \begin{minipage}{\linewidth} % 调整minipage宽度以适应公式宽度
%     \begin{equation}
%     \resizebox{\linewidth}{!}{ % 重设大小，仅调整宽度至minipage宽度
%     $\text{Score} = \frac{\text{BLEU-4} + \text{METEOR} + \text{ROUGE-L} + 0.1 \cdot \text{CIDEr}}{4} \times 100$
%     }
%     \label{eq:scale_up_bbox}
%     \end{equation}
% \end{minipage}
% \end{table}

\subsection{Implementation Details}
\textbf{Training}.
We use the pretrained LLaVA-1.6-34B as our backbone. The whole reproducible training process is only composed of the SFT stage. During the SFT stage, we focus on training the aided blocks with other parts freezing.  Table~\ref{tab: hyperparameters} shows the hyperparameters we use to finetune the CityLLaVA model. For CityLLaVA, the whole SFT phase takes 7.8 hours utilizing NVIDIA 8$\times$A100-80G GPUs.

\begin{table}[t]
    \begin{minipage}{.45\linewidth}
        \centering
        \resizebox{\textwidth}{!}{
            \begin{tabular}{c|cc}
            \hline
             Hyperparameters & Value \\
            \hline
            LR & 2e-4 \\
            Epoch & 1 \\
            BatchSize & 64 \\
            MaxLength & 2048\\
            ZeRO3 & True\\
            \hline
            \end{tabular}
        }
        \caption{The Hyperparameters for CityLLaVA.}
        \label{tab: hyperparameters}
    \end{minipage}
    \hfill
    \begin{minipage}{.45\linewidth}
        \centering
        \resizebox{\textwidth}{!}{
            \begin{tabular}{c|cc}
            \hline
            Precision   & Memory (GB) \\
            \hline
            Float16 & 78.2 \\
            INT8 & 41.8 \\
            INT4 & 22.6 \\
            \hline
            \end{tabular}
        }
        \caption{GPU resource usage during inference. The statistic is from the single GPU.}
        \label{tab:gpu_usage}
    \end{minipage}
\end{table}

\begin{table}[t]
 \resizebox{0.5\textwidth}{!}{
        \begin{tabular}{c|cccccc}
        \hline
            Model&  frames&BLEU-4 & METEOR & ROUGE-L & CIDEr & Score \\
        \hline
        Qwen-VL-Chat+BE&  8&0.243& 0.451& 0.439& 0.692& 30.03\\
        VideoLLaVA+BE&  8&0.221& 0.419& 0.426& 0.867& 28.81\\
        LLaVA-34B+LoRA&  1&0.263& 0.464& 0.455& 1.039& 32.15\\
        LLaVA-34B+BE&  1&\textbf{0.278}& \textbf{0.477}& \textbf{0.470}& \textbf{1.130}& \textbf{33.43}\\
        \hline
        \end{tabular}
    }
    \caption{The performance on different backbone and SFT method on \texttt{test} set. }
    \label{tab:backbone_sft}
\end{table}

% \begin{table}[]
%     \centering
%     \resizebox{0.2\textwidth}{!}{
%         \begin{tabular}{c|cc}
%         \hline
%          Hyperparameters & Value \\
%         \hline
%         LR & 2e-4 \\
%         Epoch & 1 \\
%         BatchSize & 64 \\
%         MaxLength & 2048\\
%         ZeRO3 & True\\
%         \hline
%         \end{tabular}
%     }
%     \caption{The Hyperparameters for CityLLaVA.}
%     \label{tab:hyperparameters}
% \end{table}

\noindent\textbf{Inference}. We use the prompts defined in Sec.~\ref{sec:text_prompt_engineering} to query the model for captions of the pedestrian and vehicle. We implement INT4 quantization for LLaVA-1.6-34B during inference. Model quantization can significantly reduce the GPU memory usage without obvious performance degradation. The statistics of GPU resource usage during inference are summarized in Table~\ref{tab:gpu_usage}. All evaluations are executed on NVIDIA 8$\times$A100-80G GPUs. The time required to evaluate the \texttt{test} set is approximately 1.7 hours. Our method takes first place in 2024 AI City Challenge Track 2 with a score of 33.4308, as shown in Table~\ref{table: ranking}. 

\begin{table}[t]
    \centering
    \resizebox{0.5\textwidth}{!}{
        \begin{tabular}{c|ccccc}
        \hline
            & BLEU-4 & METEOR & ROUGE-L & CIDEr & Score \\
        \hline
        & \multicolumn{5}{c}{View Combination (text part under single-round QA)} \\
        \hline
        Global Only  & 0.275 & 0.471 & 0.464 & 0.997 & 32.72 \\
        Local Only & \textbf{0.289} & \textbf{0.484} & \textbf{0.481} & 1.044 & 33.91 \\
        Global + Local  & 0.287 & 0.483 & 0.477 & \textbf{1.186} & \textbf{34.12} \\
        \hline
        & \multicolumn{5}{c}{Training QA manner (vision part under local view only)} \\
        \hline
        Multi-round QA & 0.252 & 0.452 & 0.442 & 0.928 & 30.94 \\
        Single-round QA & \textbf{0.289} & \textbf{0.484} & \textbf{0.481} & \textbf{1.044} & \textbf{33.91} \\
        \hline
        \end{tabular}
    }
    \caption{\textbf{The performance on different visual inputs and text inputs on \texttt{val} set}. Note that the ablation study concerning \textbf{View Combination} is carried out with text inputs in a single-round QA format, while the ablation study about the \textbf{Training QA manner} is executed with visual inputs from only a locally cropped view. \textit{Global/Local Only} refers to the performance of the model that is solely trained on global/local views. \textit{Single-round} and \textit{multi-round} \textit{QA} refer to the definition at Sec.~\ref{sec:short_qa_construction}}
    \label{tab:data_construction_ablation}
\end{table}

% \begin{table}[]
%     \centering
%     \resizebox{0.2\textwidth}{!}{
%         \begin{tabular}{c|cc}
%         \hline
%         Precision   & Memory (GB) \\
%         \hline
%         Float16 & 78.2 \\
%         INT8 & 41.8 \\
%         INT4 & 22.6 \\
%         \hline
%         \end{tabular}
%     }
%     \caption{GPU resource usage during inference. The statistic is from the single GPU.}
%     \label{tab:gpu_usage}
% \end{table}

\subsection{Ablation Study}
\label{sec:ablation_study}
We perform important ablation experiments to validate the effectiveness of the proposed modules. Note that the evaluation of LLaVA-1.6-34B requires large computational resources, we can hardly conduct the complete experiments over all different settings, and some experimental results are yielded on the \texttt{test} set rather than selected \texttt{val} set. Besides the ablation experiments about backbones and SFT methods, the other experiments are performed in LLaVA-1.6-34B equipped with block expansion.

\noindent\textbf{Effects of backbone and SFT method}. Table~\ref{tab:backbone_sft} shows the results of different backbones and SFT methods in the test set. For Qwen-VL-Chat, we simply concatenate 8 frame images as the inputs for the model training and inference. Beyond our expectations, Video-LLaVA yields the poorest score even though it was the only base model we chose that was trained with video data during both the pretraining and instruction tuning stages. TempCompass~\cite{liu2024tempcompass} also finds that most open-source video LLMs do hardly understand videos. We believe that Qwen-VL-Chat outperforms Video-LLaVa because Qwen-VL-Chat has been pretrained in more diverse and abundant data. In addition, it can be found that the increasing number of parameters can achieve better results. For LoRA and block expansion, we set the LoRA parameter of LLaVA-1.6-34b to r=256 and $\alpha$=512. Experimental results show that the block expansion performs better than LoRA.

\noindent\textbf{Effects of Prompt Engineering}. Table~\ref{tab:data_construction_ablation} shows the effects of the proposed visual and textual prompt engineering. For the ablation study about view combination,  the model based on global and local joint views outperforms models that rely solely on either global or local cues. Note that models with \textit{Global only} and \textit{Local only} are trained in \textbf{AnyRes}~\cite{liu2024llavanext} manner while yielding the suboptimal results. This comparison verifies the perspective proposed in Sec.~\ref{subsec:vpe}, unsupervised multi-cropped views contain much visual redundancy disturbing the model's attention. The local cropped view, guided by bounding boxes, provides the model with precise visual information contributing to the generation of high quality. 

\begin{figure}[t]
    \centering
    \includegraphics[width=0.5\textwidth]{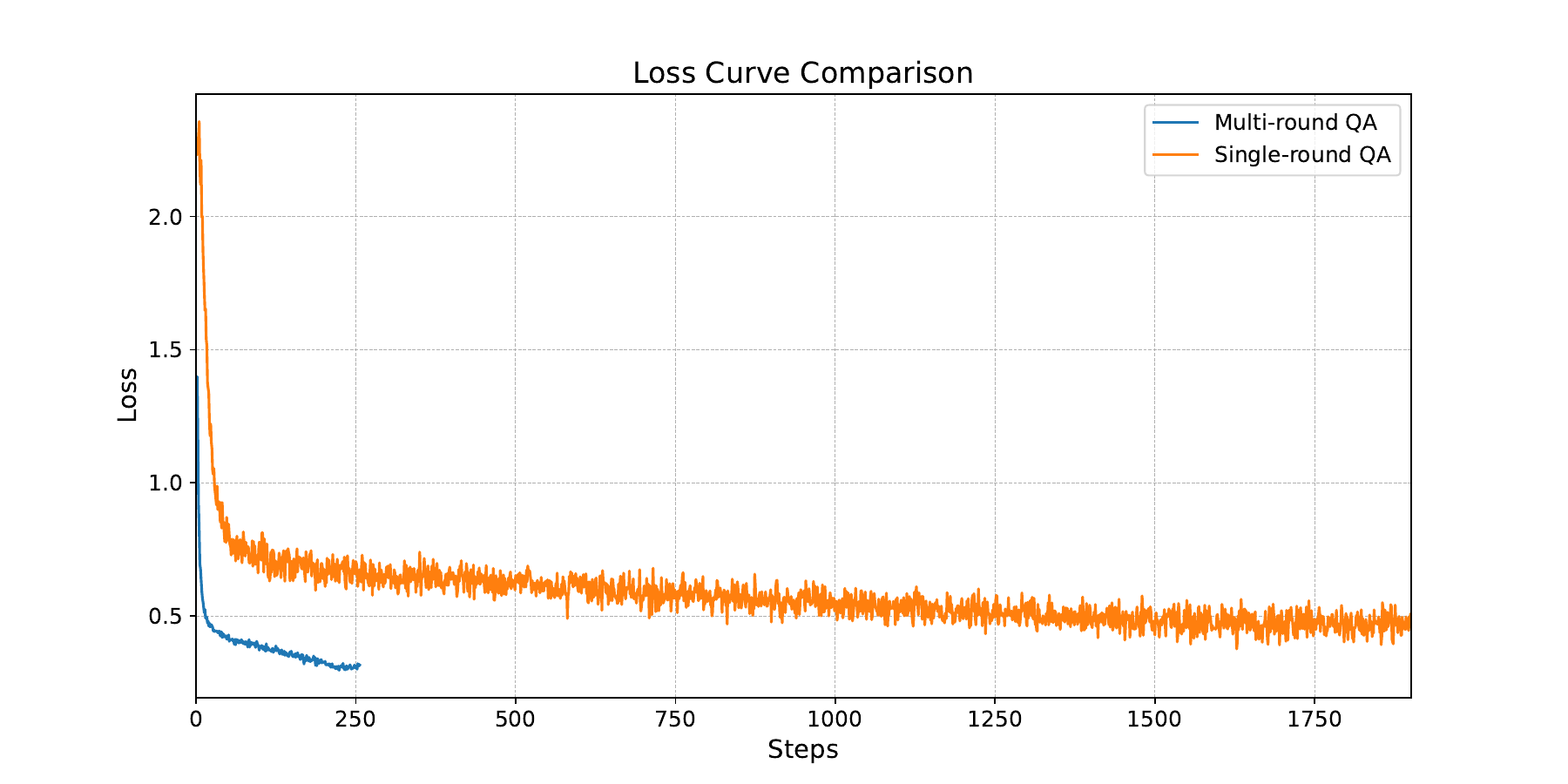}
    \caption{Training loss curves of models with multi-round and single-round QA.} 
    \label{fig:loss_curve} 
\end{figure}

The comparison between \textit{Multi-round QA} and \textit{Single-round QA} reveals that the latter format contributes to increasing the diversity of the existing dataset, an aspect that is advantageous for model training. As shown in Figure~\ref{fig:loss_curve}, the training loss for \textit{Multi-round QA} converges to approximately 0.30, whereas for \textit{Single-round QA}, it converges to around 0.45. Comprehensively considering the evaluation results and loss values, we can infer that the dataset consisting of single-round QA pairs alleviates the over-fitting during model training. 

\noindent\textbf{Effects of Sequential Questioning}. Table~\ref{tab:sequential_qa} shows the effects of different question sequences during inference. The best performance is produced with \textit{Vehicle-Pedestrian} manner. Furthermore, the caption of the pedestrian is more precise in this manner, compared to the independent QA manner. The improvement indicates that informative and precise history or prompts are beneficial to the model generation. Conversely, the hallucinatory and incorrect one prevents the model from producing high-quality responses.

\begin{table}[htb]
    \centering
    \resizebox{0.5\textwidth}{!}{
        \begin{tabular}{c|ccccc}
        \hline
            & BLEU-4 & METEOR & ROUGE-L & CIDEr & Score \\
        \hline
        & \multicolumn{5}{c}{Pedestrian Statistics} \\
        \hline
        Independent QA     & 0.214 & 0.418 & 0.360 & 0.937 & 27.14 \\
        Pedestrian-Vehicle  & 0.214 & 0.418 & 0.360 & 0.937 & 27.14 \\
        Vehicle-Pedestrian     & \textbf{0.215} & \textbf{0.425} & \textbf{0.366} & \textbf{0.989} & \textbf{27.62}  \\
        \hline
        & \multicolumn{5}{c}{Vehicle Statistics} \\
        \hline
        Independent QA     & 0.340 & 0.531 & 0.578& 1.175 & 39.16 \\
        Pedestrian-Vehicle  & 0.330 & 0.521 & 0.566 & 1.076 & 38.12 \\
        Vehicle-Pedestrian     & \textbf{0.340} & \textbf{0.531} & \textbf{0.578} & \textbf{1.175} & \textbf{39.16} \\
        \hline
        \end{tabular}
    }
    \caption{\textbf{The performance on different question sequences during inference on \texttt{val} set}. \textbf{Pedestrian-Vehicle} indicates that firstly asking \textit{pedestrian}, then asking \textit{vehicle} in a sequential conversation. Similarly, \textbf{Vehicle-Pedestrian} implies the reverse order of queries. \textbf{Independent QA} denotes separate queries for the \textit{pedestrian} and the \textit{vehicle} in distinct conversations.}
    \label{tab:sequential_qa}
\end{table}

\section{Conclusion}
This paper proposes CityLLaVA, an efficient fine-tuning for VLMs in city scenarios. Based on the modules of bounding-box guided view selection, visual prompt engineering, textual prompt engineering, short QA construction, block expansion SFT, and prediction augmentation for LLaVA. Our method obtains the best score on the leaderboard.

{
    \small
    \bibliographystyle{ieeenat_fullname}
    \bibliography{main}
}

% WARNING: do not forget to delete the supplementary pages from your submission 
% \input{sec/X_suppl}

\end{document}